\documentclass[letterpaper, 10 pt, conference]{ieeeconf}  

\IEEEoverridecommandlockouts                              

\usepackage{amsmath}
\usepackage{amsfonts}
\usepackage{amssymb}
\usepackage{paralist}
\usepackage{graphicx}
\graphicspath{ {figures/} }
\usepackage[caption=false,font=footnotesize]{subfig}
\usepackage{color}
\usepackage[ruled]{algorithm2e}
\usepackage{booktabs}
\usepackage{algpseudocode} 
\newlength\mylen
\newcommand\myinput[1]{%
	\settowidth\mylen{\KwIn{}}%
	\setlength\hangindent{\mylen}%
	\hspace*{\mylen}#1\\}
\DeclareMathOperator*{\argmin}{arg\,min}

\title{\LARGE \bf
Stochastic Functional Gradient for Motion Planning in Continuous Occupancy Maps
}

\author{Gilad Francis, Lionel Ott and Fabio Ramos{$^*$}
\thanks{$^*$ Gilad Francis, Lionel Ott and Fabio Ramos are with The School of Information Technologies, University of Sydney, Australia
        {\tt\small gfra8070@uni.sydney.edu.au}}%
}

\begin{document}

\maketitle
\thispagestyle{empty}
\pagestyle{empty}

\begin{abstract}
Safe path planning is a crucial component in autonomous robotics. The many approaches to find a collision free path can be categorically divided into trajectory optimisers and sampling-based methods. When planning using occupancy maps, the sampling-based approach is the prevalent method. The main drawback of such techniques is that the reasoning about the expected cost of a plan is limited to the search heuristic used by each method. We introduce a novel planning method based on trajectory optimisation to plan safe and efficient paths in continuous occupancy maps. We extend the expressiveness of the state-of-the-art functional gradient optimisation methods by devising a stochastic gradient update rule to optimise a path represented as a Gaussian process. This approach avoids the need to commit to a specific resolution of the path representation, whether spatial or parametric. We utilise a continuous occupancy map representation in order to define our optimisation objective, which enables fast computation of occupancy gradients. We show that this approach is essential in order to ensure convergence to the optimal path, and present results and comparisons to other planning methods in both simulation and with real laser data. The experiments demonstrate the benefits of using this technique when planning for safe and efficient paths in continuous occupancy maps.   
\end{abstract}

\section{INTRODUCTION}

Motion planning is a basic building block in autonomous robotics. Essentially, it is a decision making process that ensures safe travel from the robot's current configuration to its goal. As safety is the primary objective, the planned trajectory must avoid collision with obstacles. It is a prolific branch of robotics that has been studied for decades, producing a wide range of planning methods which can be categorically grouped into two main branches; sampling-based planning and trajectory optimisation. 
  
Planning a safe path using a \textit{Occupancy Grid Map} (OGM) is typically done by sampling-based planners \cite{Tsardoulias2016}. Most planners break the planning process into two phases. First, the planner finds a feasible, collision-free, crude path. Then, the following step improves the resulting path by applying certain heuristics. 

Trajectory optimisers optimise an objective function such as control cost or execution. However, there are no trajectory optimiser implementations for path planning using occupancy maps. The main challenge lies in the optimiser's need for contextual information anywhere along the path. Gaps or non-informative gradients will cause the optimiser to converge into a non-optimal and unsafe solution.    

In this paper, we present a new planning paradigm using occupancy maps. We utilise the recently introduced Hilbert maps \cite{ramos2015hilbert} instead of OGMs. Hilbert maps provide a fast and continuous linear discriminative model for occupancy mapping. We take advantage of the fact that spatial gradients of the occupancy can be calculated in closed form, and use them in the functional gradient motion planner update cycle. We present a novel path planner based on a \textit{Gaussian Process} (GP) path representation. Unlike other functional gradient path planning techniques (e.g. \cite{Marinho2016},\cite{Zucker2013}), the proposed planner does not commit to a predetermined resolution, whether spatial or parametric. It replaces the regularisation of the step size used in the functional gradient method with a stochastic gradient approach. This is a key element in the planner's optimisation strategy as it allows a resolution-free gradient update, which is required to ensure convergence.

The technical contributions of this paper are:
\begin{enumerate}
    \item A novel path optimisation approach for continuous occupancy maps. Effectively, this method extends previous work done on discrete cost maps to a continuous environment representation. 
	\item A stochastic functional gradient motion planner based on GP path representation. The stochastic samples allow flexible support for the path, instead of an a-priori fixed set used in previous work.
\end{enumerate}

The remainder of this paper is organised as follows. Section ~\ref{sec:RelatedWork} surveys the work related to path planning using occupancy maps. Section~\ref{sec:PathPlanning} reasons on the need for a specific planner for occupancy maps and provides details on the proposed method. Experimental results and analysis for various simulation and real data scenarios are shown in ~\ref{sec:Results}. Finally, Section~\ref{sec:conclusions} draws conclusions on the proposed method.

\section{RELATED WORK}\label{sec:RelatedWork}

Path planning using occupancy maps is commonly approached by sampling-based methods with several very successful algorithmic families such as: \textit{Rapidly exploring Random Trees} (RRT), \textit{Probabilistic RoadMap} (PRM), \textit{Visibility Graphs} (VG) and Space Skeletonisation \cite{Tsardoulias2016}. 
Sampling-based methods typically work by first building a graph representation of the configuration space, where edges represent valid connections. After the graph is built a valid path is obtained using a search algorithm on the graph structure. The visibility graphs method builds a graph where the nodes are the vertexes of the obstacles \cite{lozano1979algorithm}. Space skeletonisation uses \textit{Generalised Voronoi Diagram} (GVD) to compute safe paths \cite{bhattacharya2007voronoi, garrido2006path}. PRM randomly samples the configuration space for free space configuration and then  uses a local planner to find edges to connect these configurations to existing nodes \cite{Kavraki1996}. Next a tree search method is used to determine the path. Another successful and prolific method is RRT, which randomly grows a tree rooted at the start configuration \cite{Lavalle98rapidly-exploringrandom}. The main drawback of sampling based methods is that while they are very successful in finding safe paths, there is no explicit optimisation of an objective function, such as length or smoothness.

Optimisation is a widely used approach for finding feasible paths, where the planned path is the local extrema of a pre-defined arbitrary cost function. Loosely speaking, the cost function captures the costs and penalties associated with a configuration-space state, e.g. distance from obstacles. Khatib pioneered the use of artificial potential field for collision avoidance \cite{Khatib1986}. \textit{Covariant Hamiltonian Optimisation for Motion Planning} (CHOMP) utilises covariate gradients from a precomputed obstacle cost to minimise the trajectory's obstacle and smoothness functionals \cite{Zucker2013}. The \textit{Stochastic Trajectory Optimisation for Motion Planning} (STOMP) planner uses noisy perturbations to perform optimisation under constraints where the cost functional is non-differentiable \cite{Kalakrishnan2011}. Both CHOMP and STOMP commit to a waypoint representation which require to trade-off expressiveness with computational costs. Mukadam et al. proposed the Gaussian process motion planner which uses a Gaussian process generated by linear time varying stochastic differential equations for path representation \cite{mukadam2016gaussian}. Marinho et al. perform trajectory optimisation in a \textit{Reproducing Kernel Hilbert Space} (RKHS) \cite{Marinho2016}. However, all these methods fall short when planning using occupancy maps as discussed in section \ref{subsec:GPPaths}. 

Hilbert maps are a scalable, fast approach for continuous occupancy mapping that was recently presented in \cite{ramos2015hilbert}. The model works by a non-linear mapping of observations into a high-dimensional feature space represented by a RKHS. To ensure a linear incremental update the parameters of the model are trained by optimising a convex objective function, which ensures convergence to the global optimum. 

Traditional occupancy grid maps discretise the map into a fixed grid in order to estimate the occupancy posterior \cite{Elfes1989}. In order to make computations tractable each cell is considered as an independent random variable. The computational gains are substantial since the posterior calculation can be done separately for each cell. The drawback is the loss of spatial relationship between neighbouring cells. To alleviate this problem, a non-parametric approach based on Gaussian Processes (GPs) was proposed in \cite{OCallaghan2012}. The \textit{Gaussian processes occupancy map} (GPOM) produces probabilistic occupancy posteriors based on sensor observations. Using a parameterised covariance function, GPOM captures spatial relationships, which enables continuous inference. The computational complexity is its main limitation, as it scales cubically with the number of observations.

Hilbert maps take the advantages of both OGM and GPOM. It maps observations to a high-dimensional feature vector, whose dot product approximate the \textit{radial basis function} (RBF) kernel. Similar to GPOM, the use of kernels keeps spatial relationship which enables continuous inference. The computational complexity, on the other hand, depends linearly on the number of features. The only downside, compared with GPOM, is that posterior is a point value and not probabilistic.

\section{Path Planning using Hilbert Maps}
\label{sec:PathPlanning}
In this section we describe the proposed method, functional gradient motion planning using Hilbert maps. First, we describe the building blocks of our planner, Hilbert maps and functional gradient motion planning. Then, we present our algorithm which combines these components in a single planner.

\subsection{Hilbert Maps} \label{subsec:HilbertMaps}
In this section we briefly review the basics of Hilbert maps, which we use as our continuous occupancy representation. Building an occupancy map requires sensor (e.g. laser range finder or sonar) inputs. The dataset, $\mathcal{D}=\{\boldsymbol{x}_i,y_i\}^N_1$, contains $N$ observations, captured by the robot while moving through the environment, where $\boldsymbol{x}_i \in \mathbb{R}^D$ is a $2D$ or $3D$ position and $y_i \in \{-1,+1\}$ represents observed occupancy at $\boldsymbol{x}_i$.

The discriminative model that predicts the occupancy at a new query point, $\boldsymbol{x}^*$, is based on the \textit{logistic regression classifier} (LR) model. Given a vector of parameters $\boldsymbol{w}$ the probability of LR occupancy is given by:
\begin{equation}\label{HMAP:SimpleLR}
	p(y^* = +1|\boldsymbol{x}^*,\boldsymbol{w}) = \frac{1}{1+\exp(-\boldsymbol{w}^T\boldsymbol{x}^*)}.
\end{equation} 
As occupancy is a binary random variable, the probability of non-occupancy is given by $p(y^* = -1|\boldsymbol{x}^*,\boldsymbol{w}) = 1 - p(y^* = +1|\boldsymbol{x}^*,\boldsymbol{w})$. We note that (\ref{HMAP:SimpleLR}) is the \textit{logit} sigmoid function, $\sigma(\boldsymbol{z}_L)$ applied on the linear projection $\boldsymbol{z}_L=\boldsymbol{w}^T\boldsymbol{x}$.

The linear projection of the basic LR classifier cannot capture the complexity of a real environment. To support a richer family of functions, Hilbert maps employ nonlinear classification using approximate kernels. The kernel, $k(\cdot, \cdot)$, defines a nonlinear, and potentially infinite dimensional, mapping, $\Phi(\boldsymbol{x},\cdot)$ that projects the input into a high dimensional RKHS. The inner product between two features is then $k(\boldsymbol{x},\boldsymbol{x}') = \langle \Phi(\boldsymbol{x},\cdot), \Phi(\boldsymbol{x}',\cdot) \rangle$. To reduce model training time, Hilbert maps replace the kernel with an approximation, $\hat{\Phi}(\cdot)$ \cite{Shalev-Shwartz2011}. $\hat{\Phi}(\cdot)$ defines a finite feature vector where the dot product of these features can, in expectation, approximate the selected kernel; $k(\boldsymbol{x},\boldsymbol{x}') \approx  \hat{\Phi}(\boldsymbol{x})^T\hat{\Phi}(\boldsymbol{x}')$. Under these assumptions, the predictive occupancy posterior becomes:
\begin{equation}\label{HMAP:Prediction}
\begin{split}
	p(y^* = +1|\boldsymbol{x}^*,\boldsymbol{w}) = & 
	\frac{1}{1+\exp(-\boldsymbol{w}^T \hat{\Phi}(\boldsymbol{x}^*))} \\
	= & \sigma(\boldsymbol{z}_{NL}),
\end{split}
\end{equation}
where $\boldsymbol{z}_{NL}$ denotes the nonlinear mapping achieved by the feature vector. 
 
There are several methods to generate features to approximate a kernel \cite{ramos2015hilbert}. For the RBF kernel defined by;
\begin{equation}\label{eq:RBF}
k(\boldsymbol{x},\boldsymbol{x}') = \exp(-\gamma\parallel \boldsymbol{x}-\boldsymbol{x}'\parallel^2)
\end{equation}
there are three different approximations; Random Fourier features \cite{NIPS2008_3495}, the Nystr\"{o}m method \cite{williams2001using} and the sparse random features \cite{melkumyan2009sparse}.

After the selection of the approximation method, parameters $\boldsymbol{w}$ need to be calculated. The objective function used to estimate $\boldsymbol{w}$ is a regularised \textit{negative log-likelihood} (NLL) \cite{ramos2015hilbert}.

\subsection{Functional Gradient Motion Planning} \label{subsec:FGMP}
In this section, we describe functional gradient based optimisation methods used in path planning. 
We first introduce the notation. A path, $\xi: [0,1] \rightarrow \mathcal{C}\in\mathbb{R}^D$, is a function that maps time, $t \in [0,1]$, into configuration space $\mathcal{C}$. We define an objective functional, $\mathcal{U}(\xi): \Xi \rightarrow \mathbb{R}$, that return a real number for each path $\xi \in \Xi$. The objective functional captures the path optimisation criteria, such as the path safety and kinematic costs. 

The objective functional, $\mathcal{U}(\xi)$, varies between the different planning methods. However, it is typically a weighted sum of two penalties; \begin{inparaenum}[(i)]  
 	\item $\mathcal{U}_{obs}(\xi)$ which penalises proximity to obstacles;
 	\item $\mathcal{U}_{dyn}(\xi)$ that regulates either the curve shape or motion dynamics:
 \end{inparaenum}   
\begin{equation}\label{eq:FPMP_U}
\mathcal{U}(\xi) = \mathcal{U}_{obs}(\xi) + \lambda\mathcal{U}_{dyn}(\xi).
\end{equation}

As obstacles are defined in the robot's working space $\mathcal{W} \in \mathbb{R}^3$, estimating the obstacle cost functional is done by mapping a path from configuration space into workspace using a forward kinematic map, $x$. Given $\mathcal{B} \in \mathbb{R}^3$, a set of points on the robot, $x\left(\xi(t),u\right)$ maps a robot configuration, $\xi(t)$ and body point $u \in \mathcal{B}$ to a point in the workspace $x: \mathcal{C} \times \mathcal{B} \rightarrow \mathcal{W}$. Then, the obstacle cost functional is estimated by aggregating the workspace cost function, $c: \mathbb{R}^3 \rightarrow \mathbb{R}$, along the trajectory and robot body points using a \textit{reduce} operator such as an integral or a maximum. The only requirement is that the \textit{reduce} operator can be approximately represented by a sum over a finite set, $\mathcal{T}(\xi) = \{t,u\}_i$ of time $t_i$ and body points $u_i$:
\begin{equation}\label{eq:FGMP_Uobs}
\mathcal{U}_{obs}(\xi) \approx \sum_{(t,u)\in\mathcal{T}(\xi)} c\left(x\left(\xi(t),u\right)\right).
\end{equation}

$\mathcal{U}_{dyn}(\xi)$ is a secondary objective functional, which typically penalises based on kinematic costs or curve properties. The exact choice depends on the implementation and path representation used. In most cases the penalty deals with the derivatives of the path, for example in \cite{Zucker2013} the squared velocity norm was used as the dynamic penalty:
\begin{equation}\label{eq:velocity_norm}
	\mathcal{U}_{dyn}(\xi) = \frac{1}{2}\int_{0}^{1} \left|\left|\frac{d}{dt}\xi(t) \right|\right|^2 dt.
\end{equation}   
In \cite{Marinho2016}, the optimisation regulariser was the $L_2$ norm of $\xi$ which implicitly assumes the zero-line, connecting starting point to the goal point, as the preferable solution. We will show in section \ref{sec:Results} that such a regulariser is not suitable when planning using occupancy maps. 

Given the cost functional in (\ref{eq:FPMP_U}), optimisation of $\xi$ can be performed by an iterative approach following the functional gradient. The functional gradient update at each iteration is derived from a linear approximation of the cost functional around the current trajectory, $\xi_n$:
\begin{equation}
	\mathcal{U}(\xi) \approx \mathcal{U}(\xi_n)+\nabla_\xi\mathcal{U}(\xi_n)(\xi-\xi_n) + \mathcal{O}((\xi-\xi_n)^2).
\end{equation}
Following \cite{Zucker2013}, the optimisation update rule becomes 
\begin{equation}\label{eq:FGMP_optimisaton}
	\xi_{n+1}= \argmin_\xi \mathcal{U}(\xi_n) + (\xi-\xi_n)^T\nabla_\xi\mathcal{U}(\xi_n) + \frac{\beta}{2}\|\xi-\xi_n\|_M^2.
\end{equation}
where the term $\|\xi-\xi_n\|_M^2 = (\xi-\xi_n)^TM(\xi-\xi_n)$ is the squared norm with respect to a metric tensor $M$ and $\beta$ is a regularisation factor. A closed form solution of (\ref{eq:FGMP_optimisaton}) is obtained by differentiating the right hand side of (\ref{eq:FGMP_optimisaton}) with respect to $\xi$ and setting to zero. The update rule then becomes:
\begin{equation}\label{eq:FGMP_update_rule}
	\xi_{n+1}(\cdot)= \xi_n(\cdot) - \frac{1}{\beta}M^{-1}\nabla_\xi\mathcal{U}(\xi_n)(\cdot).
\end{equation}
Given (\ref{eq:FPMP_U}) the update rule can also be expressed as:
\begin{multline}\label{eq:FGMP_update_rule_full}
\xi_{n+1}(\cdot)= \xi_n(\cdot) - \\ \frac{1}{\beta}M^{-1}\left[\nabla_\xi\mathcal{U}_{obs}(\xi_n(\cdot)) + \lambda\nabla_\xi\mathcal{U}_{dyn}(\xi_n(\cdot))\right].
\end{multline}

  
\subsection{GP Paths using Hilbert maps} \label{subsec:GPPaths}
In this section we discuss the shortcoming of the current functional gradient methods when planning paths in occupancy maps. We then propose a method that utilises a continuous GP path representation combined with stochastic sampling to solve this problem using Hilbert maps.

\begin{figure}[thpb]
	\centering
	
	\includegraphics[width=0.3\textwidth]{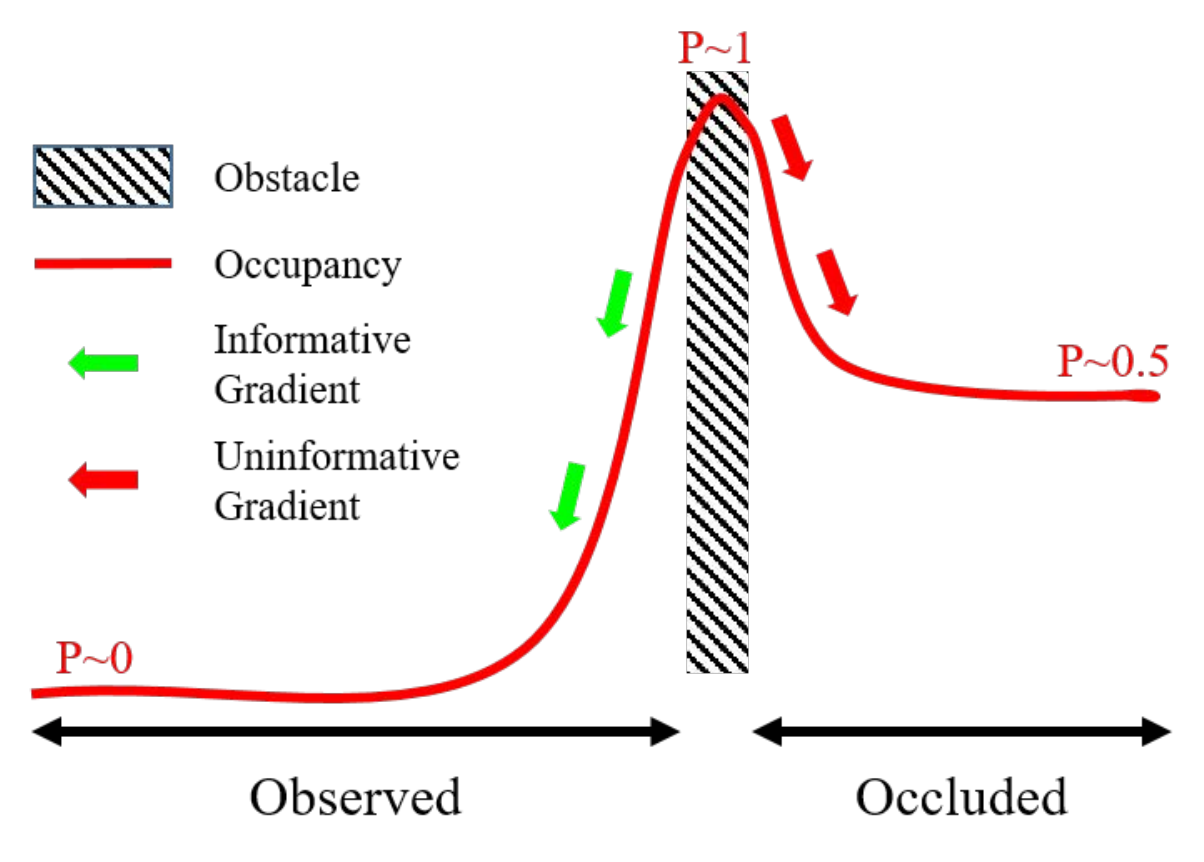}
	
	\caption{The occupancy gradient is not necessarily useful for trajectory optimisers. The occupancy, depicted in red, drops with the distance from the obstacle. However, following the occupancy gradient (green and red arrows) does not guarantee safety as it might pull (red arrows) the planner into unobserved, unsafe regions of the map.}
	\label{fig:occupancy_grad}
\end{figure}

Occupancy maps create several challenges for gradient based path planners. Most importantly, gradient information in an occupancy map is not necessarily useful. The obstacle cost functional defined in (\ref{eq:FGMP_Uobs}) requires a workspace cost function, $c(x(\xi(t),u))$. Most trajectory optimisers work with a precomputed cost map, which produce noiseless and informative gradients anywhere in the map. Estimation of the obstacle cost in an occupancy map is more challenging. While the occupancy map (or blurred map if using OGM) can act as $c(x(\xi(t),u))$, the obstacle cost is only defined well in observed areas of the map. Occluded or unreachable regions of the map are labelled as unknown with a probability of occupancy of 0.5. While the spatial gradient of occupancy "pushes" trajectory away from obstacles, the direction might be wrong as it might push the trajectory into unobserved, unsafe regions, as shown schematically in Fig. \ref{fig:occupancy_grad}.  

The other challenge arising from planning with trajectory optimisers using occupancy maps is that all planners commit to a specific path parametrisation to solve (\ref{eq:FGMP_update_rule}). Waypoint parametrisations, as used by \cite{Zucker2013, Kalakrishnan2011,park2012itomp}, have to find a balance between expressiveness and computational complexity. The more recent work in functional gradient motion planning in RKHS \cite{Marinho2016} and the Gaussian process motion planner \cite{mukadam2016gaussian,Dong2016} define a parametric support with finite resolution for their path representation. These methods produce highly expressive trajectories. However, given the finite resolution of the support, as the optimisation process deforms the trajectory, the spatial density of the support points changes. As a result some areas in the workspace have low support density, which result in a low update rate. This problem is exacerbated when using occupancy maps as some regions in the map have no informative gradients. To prevent corrupted optimisation process, such uninformative updates are rejected, reducing even further the effective density of the support.

Our approach uses Hilbert maps to produce the obstacle functional gradient. We combine a flexible stochastic gradient approach to generate support to form an expressive path based on an iterative GP representation. 

\subsubsection{Hilbert maps as cost functional}

Most functional gradient motion planners perform optimisation using a well-defined, and usually precomputed, cost map based on the distance to obstacle edges. In our approach, the spatial cost function $c(x(\xi(t),u))$ is the Hilbert occupancy map, which is estimated by (\ref{HMAP:Prediction}) along the trajectory and robot body points.
Although Hilbert maps do not form a tangible grid as OGM, querying the spatial occupancy gradient is as straightforward as querying and occupancy grid. By applying the chain rule, the Euclidean space gradient of (\ref{HMAP:Prediction}) around a query point, $\boldsymbol{x}^*$ becomes:
\begin{equation}\label{eq:HMAP:grad}
\begin{split}
	\frac{\partial}{\partial\boldsymbol{x}^*}p(y^* = +1|\boldsymbol{x}^*,\boldsymbol{w})=\frac{\partial \sigma(\boldsymbol{z_{NL}})}{\partial \boldsymbol{z}_{NL}}\frac{\partial \boldsymbol{z}_{NL}}{\partial \boldsymbol{x}^*}  \\
	\approx \sigma(\boldsymbol{z}_{NL})(1-\sigma(\boldsymbol{z}_{NL}))\boldsymbol{w}^T\frac{\partial }{\partial\boldsymbol{x}^*} \hat{\Phi}(\boldsymbol{x}^*).	
\end{split}
\end{equation}
Here we used the fact the the derivative of LR is given by $\frac{\partial}{\partial \boldsymbol{x}}\sigma(\boldsymbol{x})=\sigma(\boldsymbol{x})(1-\sigma(\boldsymbol{x}))$.


\subsubsection{GP Path}

GPs provide a principled way to represent smooth trajectories. The GP model requires a small set of support points, which define waypoints in configuration space the path should follow. GP regression provides us with a complete solution that allows querying the model at any given time and handles boundary conditions we wish to impose. Unlike other GP-based motion planners \cite{mukadam2016gaussian,Dong2016}, the method presented here does not require a predefined support, but rather learns and builds the trajectory support while optimising the path.

A GP path is defined as a vector-valued (multiple output) GP \cite{Alvarez2011}: 
\begin{equation}
\xi(t) \sim \mathcal{GP}(\mu(t), {K}(t,t')), \qquad t,t' \in [0,1].
\end{equation}   
Here, $\mu(t) \in \mathbb{R}^D$ is the vector-valued mean function of $t$ and ${K}(t,t') \in \mathbb{R}^D \times \mathbb{R}^D$ is a positive matrix-valued kernel between $\xi(t)$ and $\xi(t')$ with a corresponding kernel matrix for two different time instances, $\boldsymbol{K}(t,t')$:
\begin{equation}
\boldsymbol{K}(t,t')=
\begin{bmatrix}
k_{1,1}(t,t') & k_{1,2}(t,t') & \dots & k_{1,D}(t,t')\\
\vdots & \ddots &  & \vdots\\
k_{D,1}(t,t') & k_{D,2}(t,t') & \dots & k_{D,D}(t,t').
\end{bmatrix}
\end{equation}
Each element in $\boldsymbol{K}$, $k_{d,d'}(t,t')$ represents the effect joint $[\xi{t}]_d$ at time $t$ has on joint $[\xi{t'}]_{d'}$ at $t'$.

Updating the model requires conditioning the GP model with waypoint observations. The term observations here is used loosely, and means the states, $\xi^o(t^o)$ at time $t^o$ that the trajectory must pass through. By conditioning the GP with these observations we can compute the \textit{maximum a posteriori} (MAP) path at any query time, $t^*$:
\begin{equation}\label{eq:GP_posterior}
	\bar{\xi}(t^*) = \mu(t^*) + \boldsymbol{K}(t^*,t^o)\boldsymbol{K}(t^o,t^o)^{-1}(\xi^o(t^o)-\mu(t^o)).
\end{equation}
The choice of $t^o$ differentiates this work from other GP path representations, such as \cite{mukadam2016gaussian,Dong2016}. While in previous work $t^o$ was fixed a-priori, in this work $t^o$ is learned online. 

There are several advantages of using GPs to represent the path. First, we do not need to discretise the path. Instead a finite set of $N$ points, $\{t_i^o, \xi_i^o(t_i^o)\}_{i=1}^N$, serve as the curve support, which can be queried for its MAP value at any given time $t^*$ using (\ref{eq:GP_posterior}). Second, the mean function $\mu$ provides an explicit prior, which can be exploited by initialising the optimisation with a rough path from a fast path planning method. Finally, boundary conditions can be imposed by treating them explicitly as observations. Besides the obvious boundary conditions at the start and goal points one can define must-visit waypoints along the trajectory or define the robot's direction by including derivative observations \cite{Solak2002}.

\subsubsection{Stochastic Gradient}

A drawback of other functional gradient motion planners is that they either utilise a spatial parameterisation or commit to a finite parametric resolution to represent and update the path. In both cases, this may lead to gaps in the sampling of the objective functional. To overcome this we adopt a resolution-free sampling method. Since the functional objective of (\ref{eq:FPMP_U}) can be approximated by a sum of individual points along the path, optimisation of the objective can be performed using \textit{stochastic gradient descent} (SGD) \cite{Bottou2010}. From (\ref{eq:FPMP_U})-(\ref{eq:velocity_norm}), we define an empirical objective functional that approximates the real objective:
\begin{equation}\label{eq:empirical_risk}
	\hat{\mathcal{U}}(\xi) = \sum_{t,u} \mathcal{U}_{obs}\left(\xi(t,u)\right) + \lambda\mathcal{U}_{dyn}\left(\xi(t,u)\right) \xrightarrow[n \rightarrow \infty]{} \mathcal{U}(\xi).
\end{equation}

A consequence of (\ref{eq:empirical_risk}) is that minimising the objective requires reasoning over many points all along the curve. Such a process, which effectively resembles batch optimisation, is computationally infeasible. The approach taken by other trajectory optimisation methods (e.g. \cite{Marinho2016,mukadam2016gaussian}) is to estimate $\hat{\mathcal{U}}(\xi)$ with a finite resolution support. However, it is clear from (\ref{eq:empirical_risk}) that while this is computationally attractive, such an approach cannot guarantee convergence to the optimum.

The stochastic functional gradient path planner utilises SGD to ensure convergence \cite{bottou2016optimization}. SGD randomly selects a mini-batch from the dataset and then updates the solution in small steps, based on the gradient computed from that mini-batch. In a similar fashion, our method uses random samples $[t^*, u^* \xi_n(t^*)]$ as stochastic training points. The update rule of \ref{eq:FGMP_update_rule} is then replaced by:   
\begin{equation}\label{eq:stochastic_update_rule}
\xi_{n+1}(\cdot)= \xi_n(\cdot) - \eta_n A^{-1}\nabla_\xi\mathcal{U}(\xi_n(t^*),u^*)
\end{equation}
where $\eta_n > 0$ is the step size parameter and can be either constant or asymptotically decaying. Matrix $A$ can be seen as a preconditioner that may help accelerate convergence rate, and in many cases is set to the identity matrix \cite{bottou2016optimization}. Note the difference between (\ref{eq:FGMP_update_rule}) and (\ref{eq:stochastic_update_rule}) where the constant regulariser $\beta$ is replaced by $\frac{1}{\eta_n}$.

Using SGD ensures the convergence, in expectation, of the empirical objective in (\ref{eq:empirical_risk}) to the optimal objective, while keeping computational cost per iteration low \cite{bottou2016optimization}.   
\subsubsection{Planning on Hilbert Maps Algorithm}
 
To finalise the stochastic functional gradient path planning algorithm, we need to define the functional gradient of $\mathcal{U}_{obs}$ and $\mathcal{U}_{dyn}$. For a general functional of the form $\mathcal{F}(\xi)=\int_{a}^{b} v(t,\xi,\xi') dt$, the gradient is given by:
\begin{equation}\label{eq:functional_grad}
	\nabla\mathcal{F}_\xi(\xi)=\frac{\partial v}{\partial\xi} - \frac{d}{dt}\frac{\partial v}{\partial\xi'}.
\end{equation}

Applying (\ref{eq:functional_grad}) to $\mathcal{U}_{obs}$ at a sampled time $t^*$ and for robot body point $u^*$ yields;
\begin{equation}\label{eq:d_U_obs-final}
\nabla\mathcal{U}_{obs}(\xi(t^*),u^*)=\frac{\partial}{\partial \xi(t)}x(\xi(t^*),u^*) \nabla_x c\left(x\left(\xi(t^*),u^*\right)\right).
\end{equation}
Here, $\boldsymbol{J}(t^*,u^*) \equiv \frac{\partial}{\partial \xi(t)}x(\xi(t^*),u^*)$ is the workspace Jacobian. $\nabla_x$ emphasises that this is a Euclidean gradient of the cost function $c$.

We opted to use the squared velocity norm integral, as shown in (\ref{eq:velocity_norm}), as the dynamic penalty $\mathcal{U}_{dyn}$. Using the $L_2$ norm such as in \cite{Marinho2016} is less attractive as it implicitly defines a favourable simple mean solution which requires tuning of regularisation coefficients for different scenarios. With (\ref{eq:velocity_norm}), the functional gradient can be easily computed as:
\begin{equation}\label{eq:d_U_reg-final}
	\nabla\mathcal{U}_{dyn}(\xi(t^*))= -\frac{d^2}{dt^2}\xi(t^*).
\end{equation}
Again, as $\xi(t)$ is represented by a GP, computing the derivative is straight-forward.  
The update rule at time $t^*$ and robot body point $u^*$ can now be summarised from (\ref{eq:stochastic_update_rule}), (\ref{eq:d_U_obs-final}), and (\ref{eq:d_U_reg-final}) as:
\begin{multline}\label{eq:final_update_rule}
\xi_{n+1}(t^*)= {\xi}_n(t^*) - \\
\eta_n A^{-1}\left(\boldsymbol{J}(t^*,u^*)^T\nabla_x c\left(x\left({\xi}(t^*),u^*\right)\right) + \lambda\frac{d^2}{dt^2}{\xi}(t^*)    \right).
\end{multline}

\begin{algorithm}[bt]
	\caption{Functional gradient path planning using Hilbert maps}
	\label{algo:main}
	\DontPrintSemicolon
	\KwIn{$\mathcal{H}$: Hilbert Occupancy Map.}
	\myinput{$\xi(0),\xi(1)$: Start and Goal states.}
	\myinput{$P_{safe}$: No obstacle Threshold.}
	\myinput{$\xi_{initial}(t)$: Initial solution (optional).}
	\myinput{${K}$: covariance function for GP path.}
	\KwOut{$\xi_{min}(t)$}
	
	\tcp{Use prior guess/solution if available}
	\eIf{$\xi_{initial}$}
	{$\mu_0\leftarrow \xi_{initial}$}
	{$\mu_0\leftarrow \left(\xi(1)-\xi(0)\right)t + \xi(0)$}
	
	$\xi_0 \leftarrow \mathcal{GP}_0\sim \mathcal{GP}(\mu_0,{K})$\\
	$n=0$\\
	\While{$\xi$ not converged}
	{
		\tcp{Stochastic sampling}
		$(t_{sup},u_{sup}) \leftarrow$ Draw mini-batch randomly\\
		\ForEach{$(t^*,u^*) \in (t_{sup},u_{sup})$}
		{$P_{occ} \leftarrow \mathcal{H}(x(\xi_n(t^*),u^*))$ Eq. (\ref{HMAP:Prediction})\\
			\If{$P_{occ} \leq P_{Safe}$}
			{$\xi_{n+1}(t^*) \leftarrow$ update rule Eq. \ref{eq:final_update_rule}\\
				
				$\xi_{n}(t) \leftarrow \text{update GP with }(t^*, \xi_{n+1}(t^*))$	
			}
		}
		
		\tcp{Update boundary conditions}
		$\xi_n(t) \leftarrow \text{update: } (0, \xi(0)),(1, \xi(1))$
	
		$\xi_{n+1}(t) \leftarrow \mathcal{GP}(\xi_n(t), {K})$ \\
		
		$n=n+1$	
	}
	
\end{algorithm} 

The algorithm for path planning using Hilbert maps is shown in Algorithm (\ref{algo:main}). The algorithm accepts an optional initial solution to start optimisation with, otherwise a straight line trajectory is used initially. This initial path is then used as the mean function of the GP path. 
   
At each iteration, a mini-batch $(t_{sup}, u_{sup})$ is drawn randomly. For each point $t^* \in t_{sup}$ and $u^* \in u_{sup}$ the corresponding state $\xi_{n}(t^*)$ is computed from the active GP path model, $\xi_{n}(t)$. To perform the functional update, the Hilbert map is queried at $x(\xi(t^*),u^*)$, and the probability of occupancy $P_{occ}$ is obtained. Functional updates may only happen if the occupancy is within safe limits, i.e. free from obstacles. Using (\ref{eq:final_update_rule}) new states $\xi_{n+1}(t^*)$ are computed and the path model $\xi_n(t)$ is updated with the new path observations. To enforce a valid path the boundary conditions are then incorporated into the GP model as additional observations. Finally, a new path model is initialised with the previous model as its mean function ${\xi}_{n+1}(t) \sim \mathcal{GP}({\xi}_n(t),{K})$.  

\section{Results}\label{sec:Results}
In this section, we evaluate the performance of the stochastic functional gradient path planner in simulation and with real data and provide comparisons to other methods.

\subsection{Simulations}\label{subsec:simulation}

In this section we compare the proposed method with the functional gradient motion planner in RKHS \cite{Marinho2016}, as both methods are related. We show that while \cite{Marinho2016} provides a flexible path representation, changes are needed in order to perform optimisation in occupancy maps. 

\begin{figure}[!ht]
 \centering
 	\subfloat[Optimisation Process \label{RKHS:subfig-1}]{%
 		\includegraphics[width=0.14\textwidth]{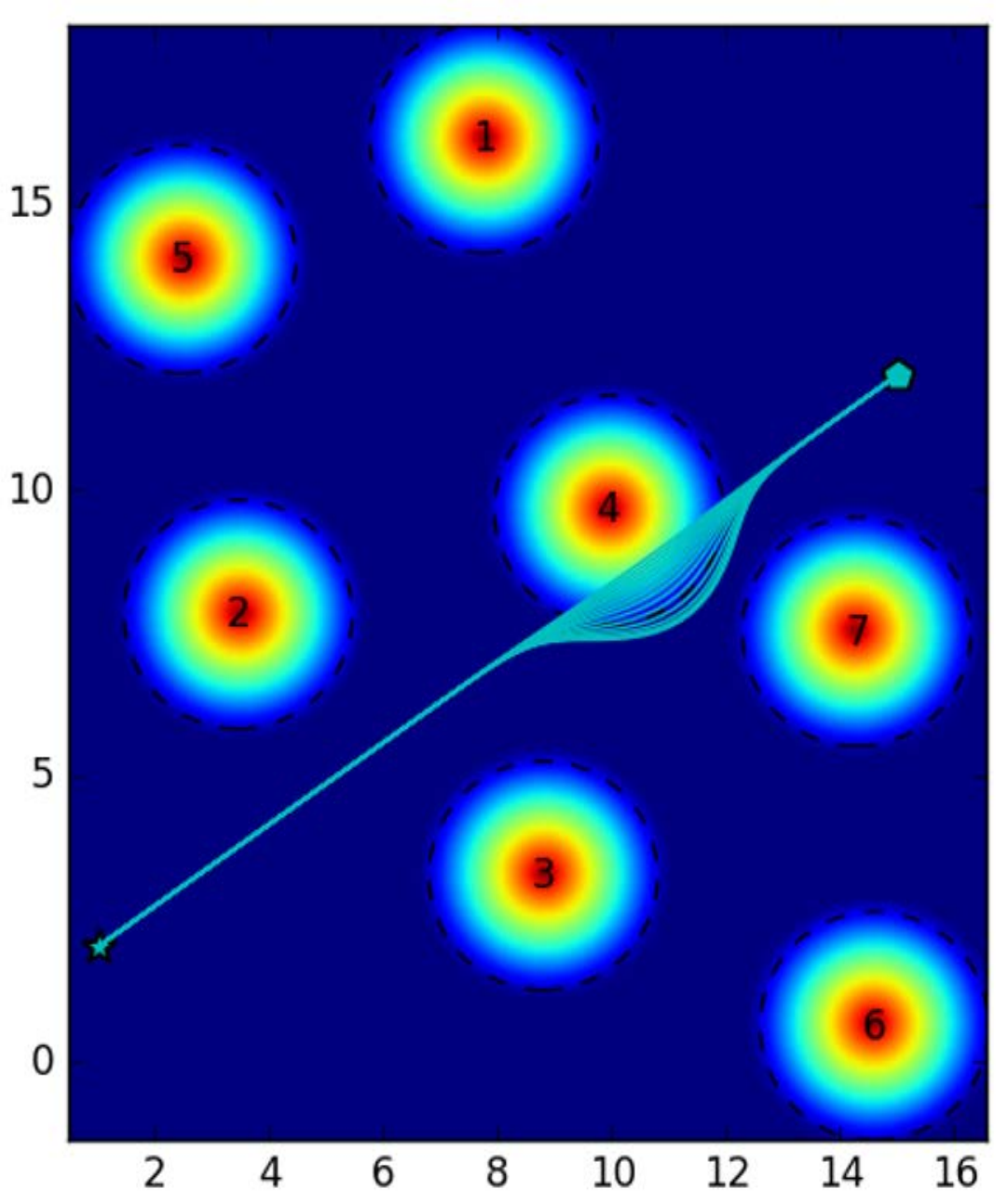}
 	}
 	\hspace{15pt}
 	\subfloat[Final Path\label{RKHS:subfig-2}]{%
 		\includegraphics[width=0.14\textwidth]{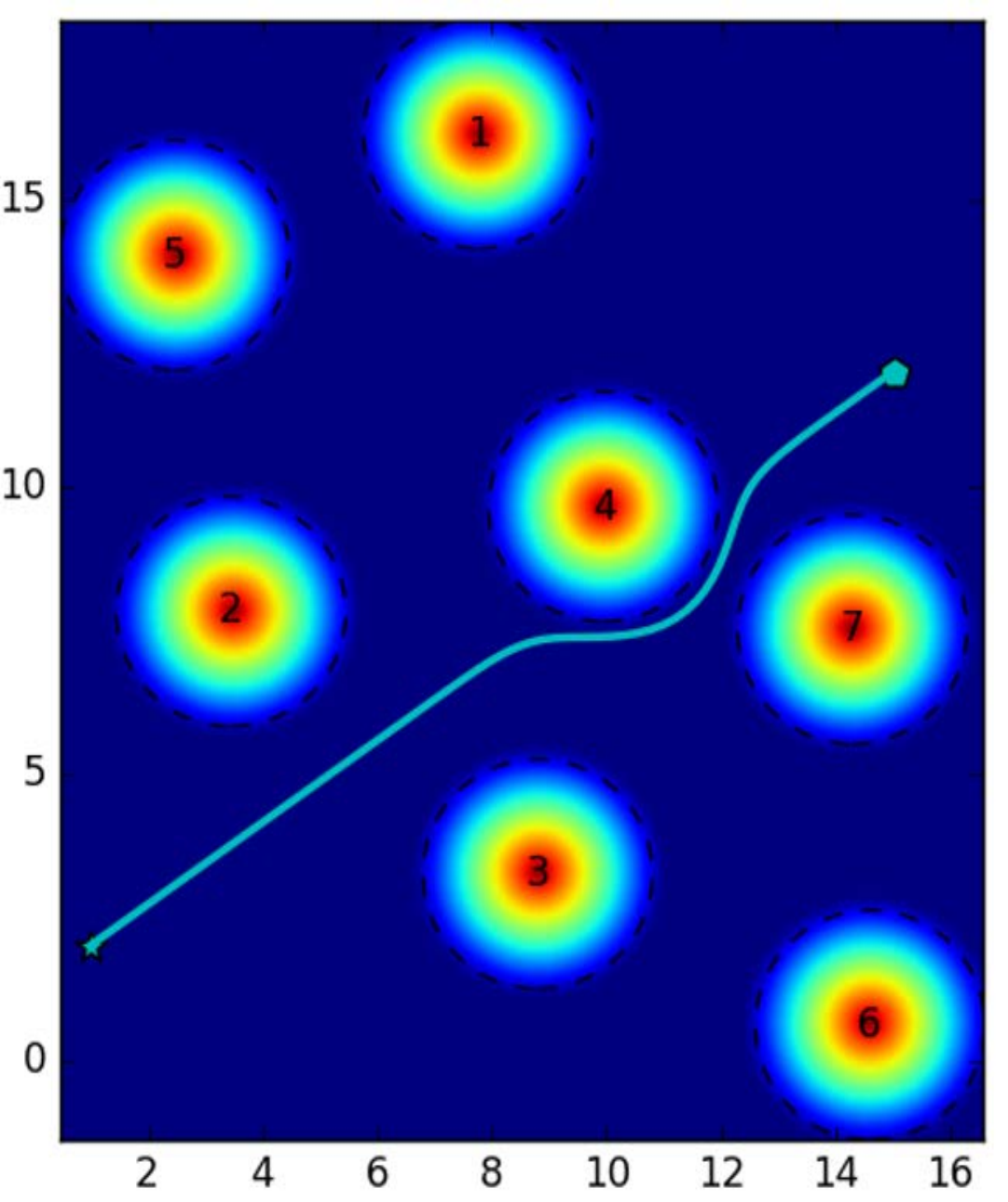}
 	}
 	\caption{RKHS motion planning \cite{Marinho2016} in a precomputed cost field calculated according to (\ref{eq:distance_cost}). Obstacle potential costs and their spatial gradients are well-defined anywhere in the workspace. Dashed lines illustrate the obstacles boundaries. Star and pentagon mark the start and goal points, respectively. \protect\subref{RKHS:subfig-1}  paths generated during the RKHS optimisation process. \protect\subref{RKHS:subfig-2} shows the optimal path.  }
 	\label{fig:RKHS}
 \end{figure}

Most trajectory optimisers assume full knowledge about the location of obstacles and precompute offline a cost field, $c(\boldsymbol{x})$ in workspace $\mathcal{W}$ which penalises the proximity to obstacles, for example \cite{ratliff2009chomp}:
\begin{equation}\label{eq:distance_cost}
c({\boldsymbol{x}})=
\begin{cases}
	-d(\boldsymbol{x})+\frac{1}{2}\epsilon & \text{ $d(\boldsymbol{x}) < 0$}\\
	\frac{1}{2\epsilon}(d(\boldsymbol{x})-\epsilon)^2 & \text{$ 0 \leq d(\boldsymbol{x}) \leq \epsilon$}\\
	0 & \text{otherwise}\\
\end{cases},\\
\end{equation}      
where $d(\boldsymbol{x})$ is the distance of $\boldsymbol{x}$ to the boundary of the nearest obstacle and $\epsilon$ is a minimal safety buffer from obstacles. An example of path planning with a precomputed cost field based on \cite{Marinho2016} is shown in Fig. \ref{fig:RKHS}: \ref{RKHS:subfig-1} depicts the iterative optimisation process and \ref{RKHS:subfig-2} shows the optimal path. These noiseless obstacles result in a cost gradient that is well-defined anywhere in the workspace, including inside obstacles. As the gradient is precomputed, evaluating the update rule is also computationally efficient, leading to fast convergence at a local minima. However, planning using occupancy maps adds several challenges to this optimisation process. The cost field, and more importantly its spatial gradients, are not necessarily informative. In addition, as the map is generated by laser observations, all predictions are noisy. As a result, the assumptions at the core of the functional gradient-based planner are no longer valid and a change to the planning process is required.    

\begin{figure}[thpb]
	
	\centering
	
	\includegraphics[width=0.15\textwidth]{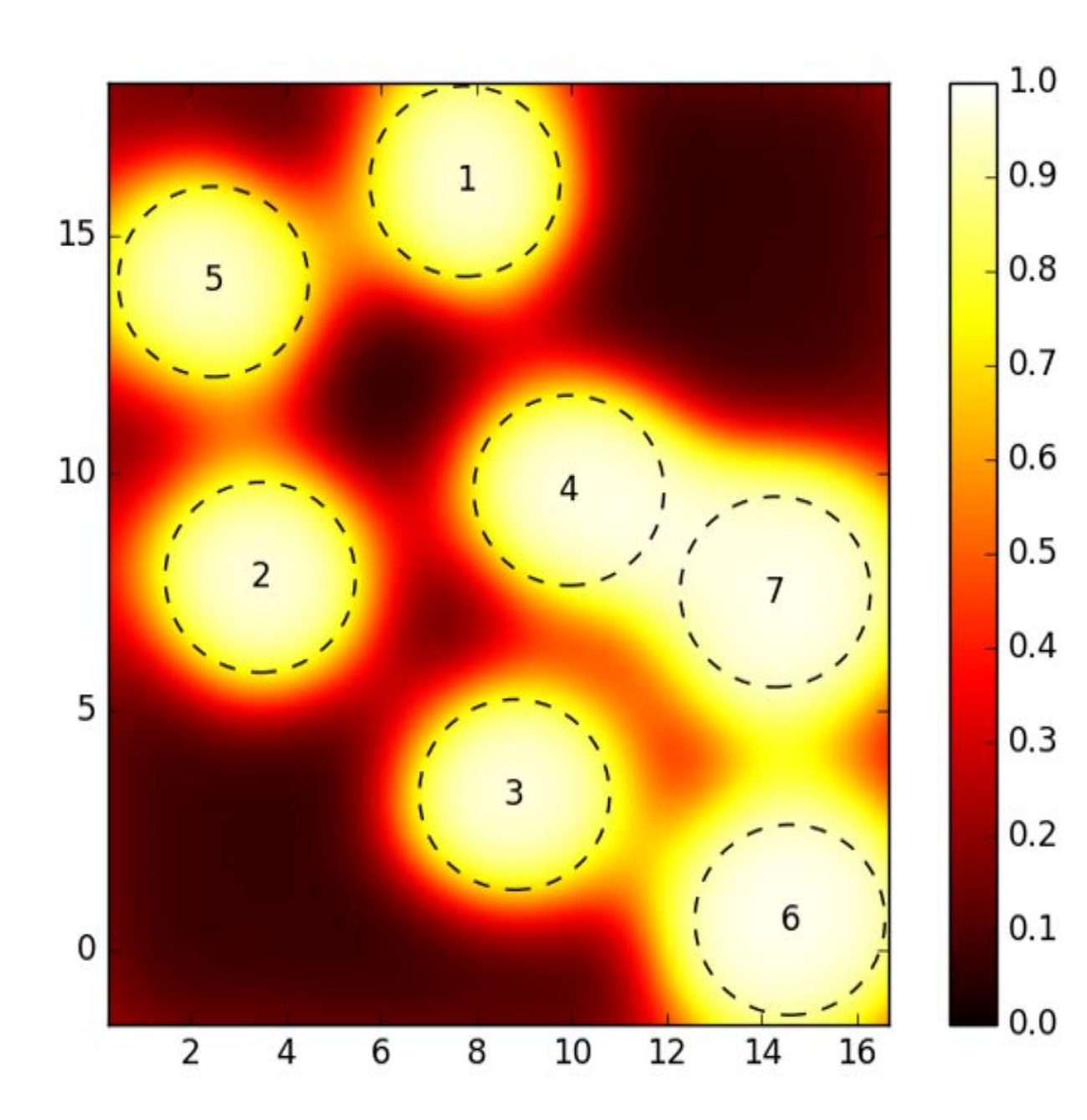}
	
	\caption{Continuous occupancy Hilbert map for the environment shown in Fig \ref{fig:RKHS}. The map shows the probability of occupancy, $p(y^* = +1|\boldsymbol{x}^*,\boldsymbol{w})$ as in (\ref{HMAP:Prediction}). Note that the optimal path of in Fig. \ref{fig:RKHS} passes through the gap between obstacles 4 and 7. In the Hilbert map, such a path is considered invalid as it passes occupied or unsafe area.}
	\label{fig:hmap}
    \hspace{-5mm}
\end{figure}

Constructing a Hilbert map for a simulated environment of randomly placed obstacles is achieved by generating an occupancy dataset. The dataset  $\mathcal{D}=\{\boldsymbol{x}_i,y_i\}^N_1$ is created by randomly placing occupied observations, $y_i = +1$,  on the boundaries of all obstacles and free observations $y_i = -1$ outside obstacles. There are no laser points inside obstacles. After the dataset is created, the map model is fitted. Fig. \ref{fig:hmap} depicts a Hilbert continuous occupancy map generated for the environment shown is Fig \ref{fig:RKHS}.
Using the continuous map representation, the occupancy and its gradient can be queried at any location. 

\begin{figure}[thpb]
	
	\centering
	
	\includegraphics[width=0.35\textwidth]{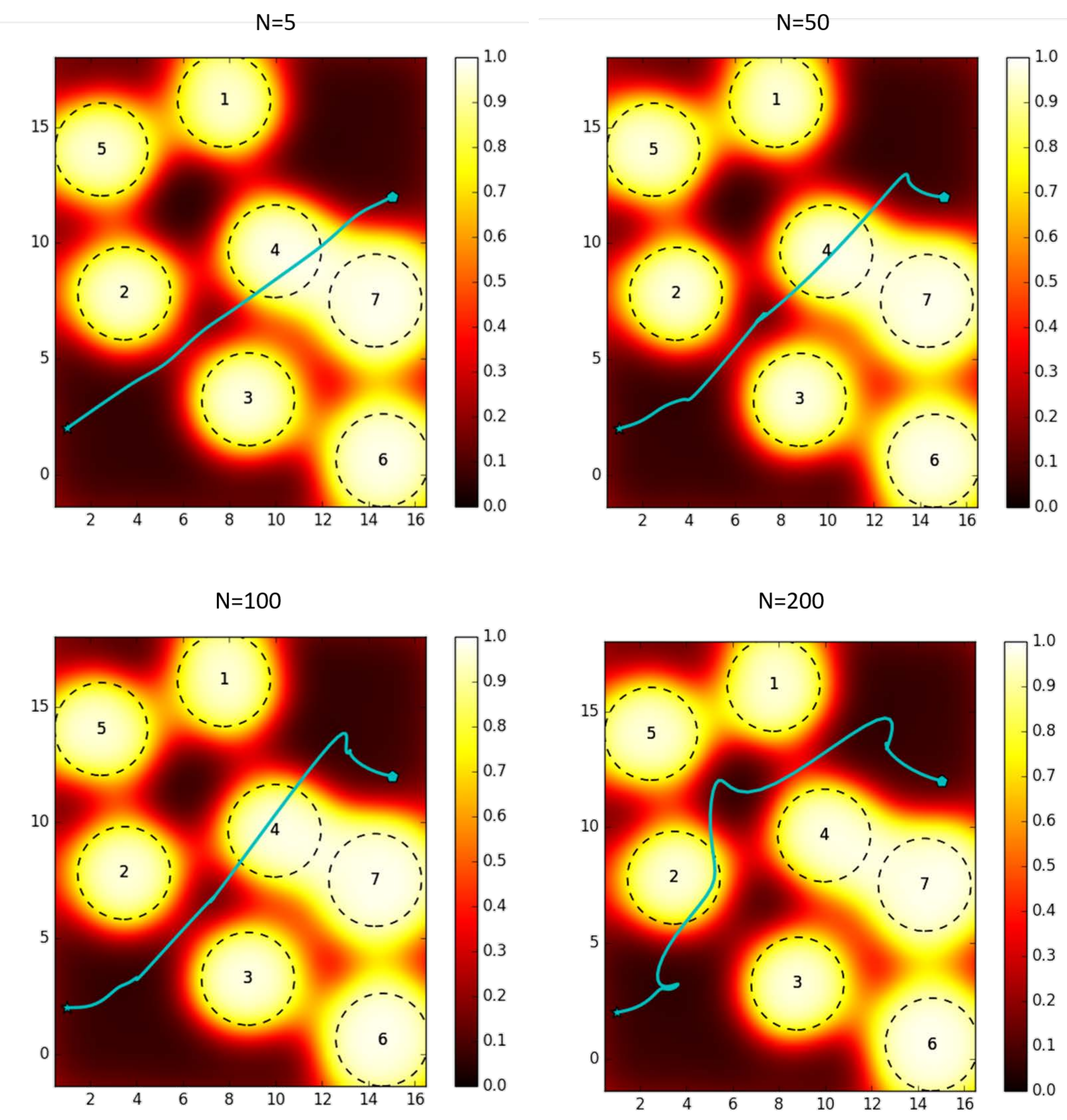}
	
	\caption{RKHS motion planning \cite{Marinho2016} fails to plan using Hilbert maps. The individual plots compare the effect the size of the path support $N$ has. The final path is depicted in cyan, while the cyan star and pentagon mark the start and goal points, respectively. The RKHS motion planner does not find a valid solution since it represents the path with a parametric resolution, which leads to gaps in the sampling of the objective functional.}
	\label{fig:RKHS_hmap_compare}
\end{figure}

Fig. \ref{fig:RKHS_hmap_compare} shows an attempt to plan a path in an occupancy map using the functional gradient method described in \cite{Marinho2016} with various support size ($N=5,50,100,200$). The cyan line depicts the optimal path after the algorithm has converged. Clearly, the resulting path is unsafe regardless of the size of support, $N$, used in the optimisation process. As expected, increasing the size of support lead to a more expressive path. Yet, even with $N=200$ the resulting path was not safe. The reason lies in the lack of informative gradient in the occupancy map and the finite parametric resolution of the path representation. As the optimisation process deforms the curve, the spatial density of the support changes too.  Consequently, there is less support around critical areas of the map such as the boundaries of obstacles. Increasing resolution even further will not alleviate this problem since we have no \textit{a-priori} knowledge of the number of required support points. Inadvertently, increasing the number of support points create unnecessary jerks in the curve, as a response to noisy occupancy gradient, as shown in Fig \ref{fig:RKHS_hmap_compare}.

\begin{figure}[!ht]
 \centering
 	\subfloat[Average optimal path \label{GPPlanner_simulation:subfig-1}]{%
 		\includegraphics[width=0.25\textwidth]{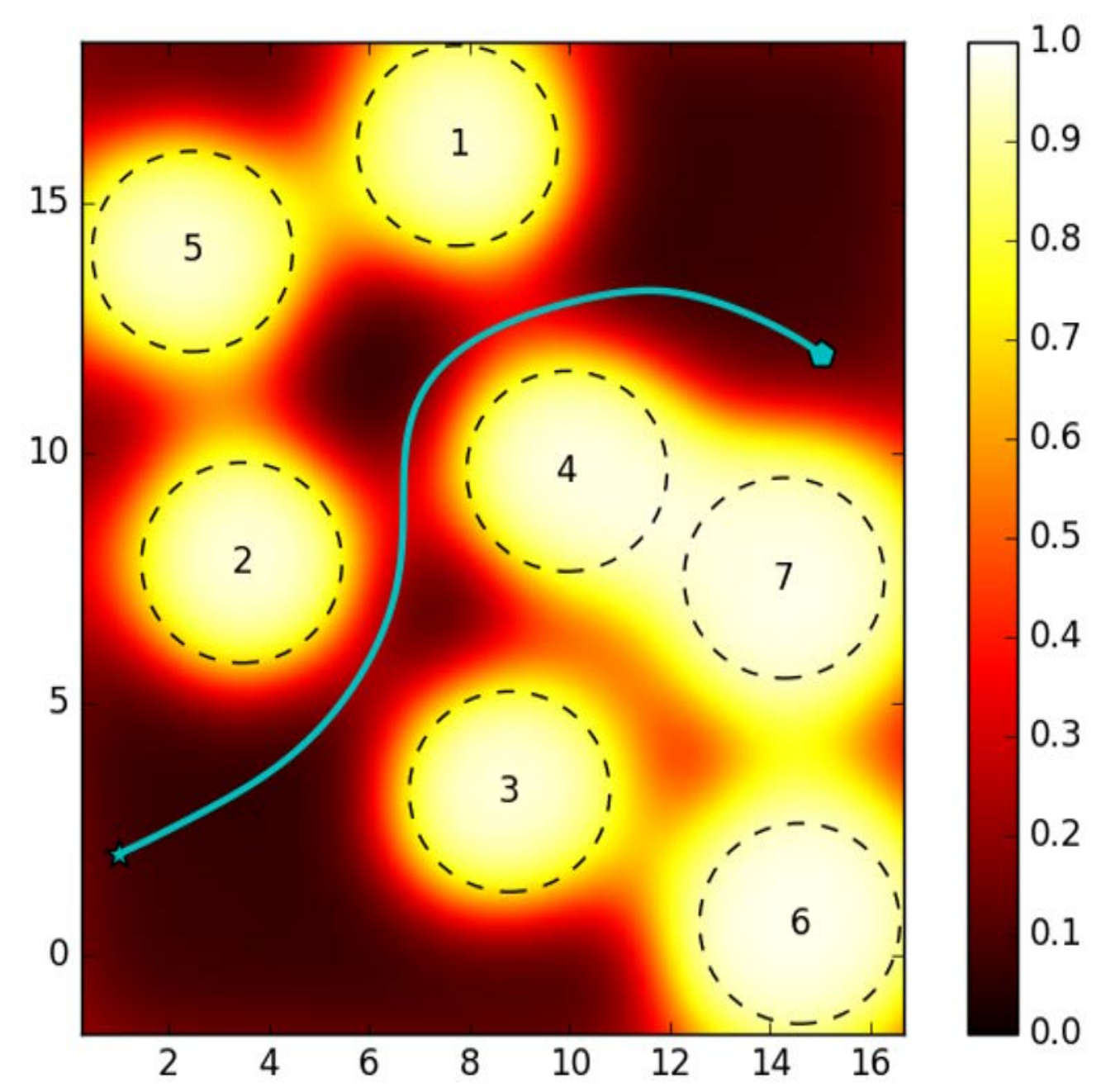}
 	}
 	\hspace{5pt}
 	\subfloat[Convergence \label{GPPlanner_simulation:subfig-2}]{%
 		\includegraphics[width=0.3\textwidth]{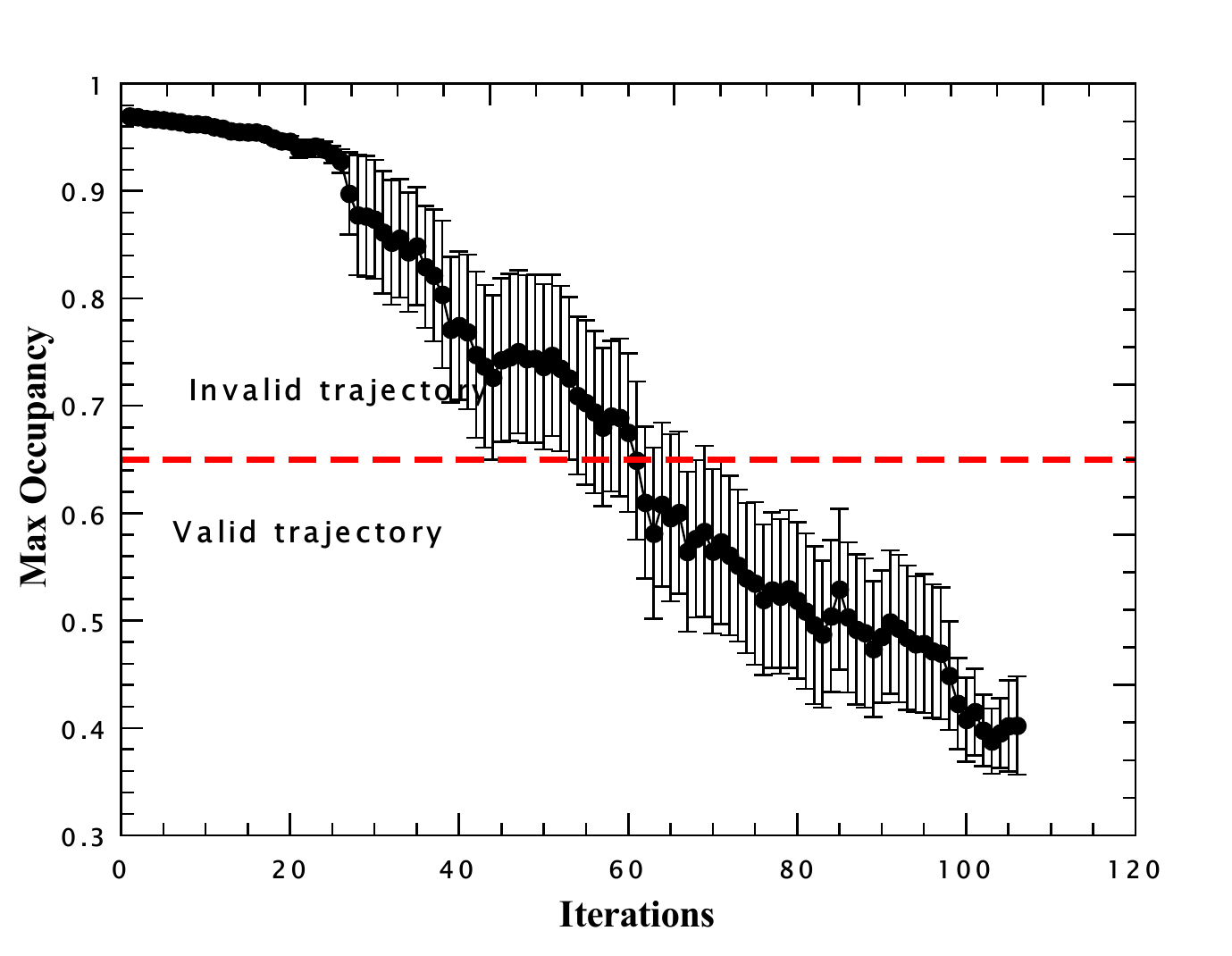}
 	}
 	\caption{Stochastic functional gradient motion planner results. \protect\subref{GPPlanner_simulation:subfig-1} Average path over 10 repetitions. The trajectory is depicted in cyan while the star and pentagon marker indicate the start and goal points, respectively. The average path follows the mid line between obstacles, which reduces the obstacle cost. \protect\subref{GPPlanner_simulation:subfig-2} Shows the convergence of the stochastic functional gradient motion planner. The maximum occupancy along the trajectory is plotted as a function of the iterations. Data shown is the average over 10 repetitions. The $P=0.5$ dashed red line marks the threshold for a valid trajectory. Note that the maximum occupancy does not reduce to zero, as the continuous occupancy map predictions for the gap between obstacles 1,2 and 4 are approximately 0.4.  }
 	\label{fig:GPPlanner_simulation}
 \end{figure}

Our stochastic functional gradient method does not commit to a specific parametric resolution, rather it randomly selects points along the curve during the optimisation. Fig. \ref{GPPlanner_simulation:subfig-1} shows in cyan the average path planned using the occupancy map of Fig \ref{fig:hmap}. Since the optimisation objective balances an obstacle cost with a penalty on the trajectory shape, the optimal path is collision-free and smooth. Finding a path relies on generating enough samples to instantiate a gradient update, especially in areas of high importance such as obstacles boundaries. Fig. \ref{GPPlanner_simulation:subfig-2} shows the convergence of the optimisation process to the minima of the objective, by plotting the maximum occupancy along the trajectory at each iteration. The maximum occupancy along the path drops steadily as the optimisation progresses. However, it cannot drop further than 0.4, since the predictive occupancy between obstacles 2 and 4 is approximately that number. Yet, Fig. \ref{GPPlanner_simulation:subfig-2} provides empirical evidence for the expected optimality of the stochastic process.       

\subsection{Real data}\label{subsec:real_data}

\begin{figure}[!ht]
	\centering
 	
    \subfloat[SFGMP \label{Intel_Comparison:subfig-1}]{%
 		\includegraphics[width=0.23\textwidth]{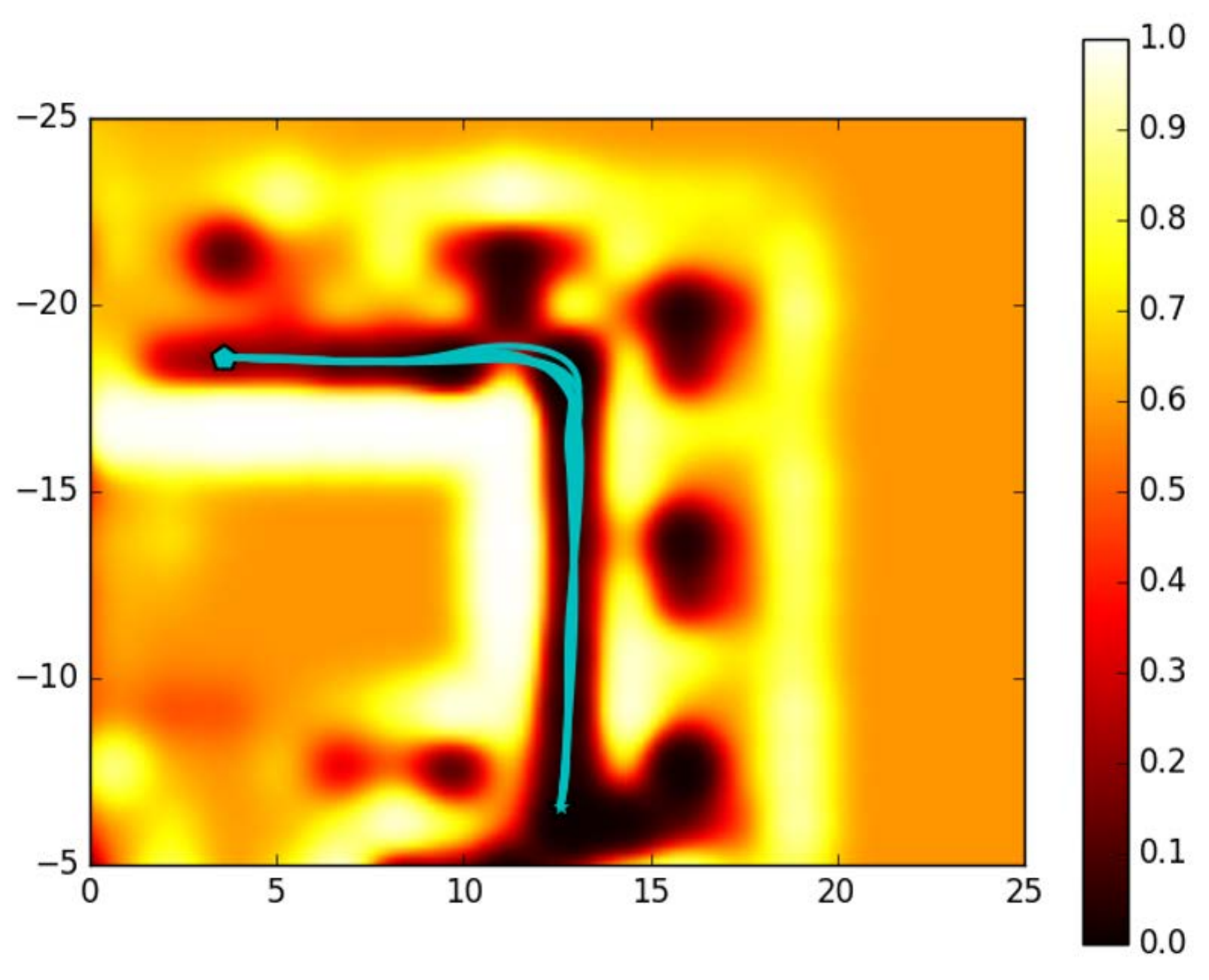}
 	}
 	\hspace{1pt}
 	\subfloat[$\text{RRT}^*$ \label{Intel_Comparison:subfig-2}]{%
 		\includegraphics[width=0.23\textwidth]{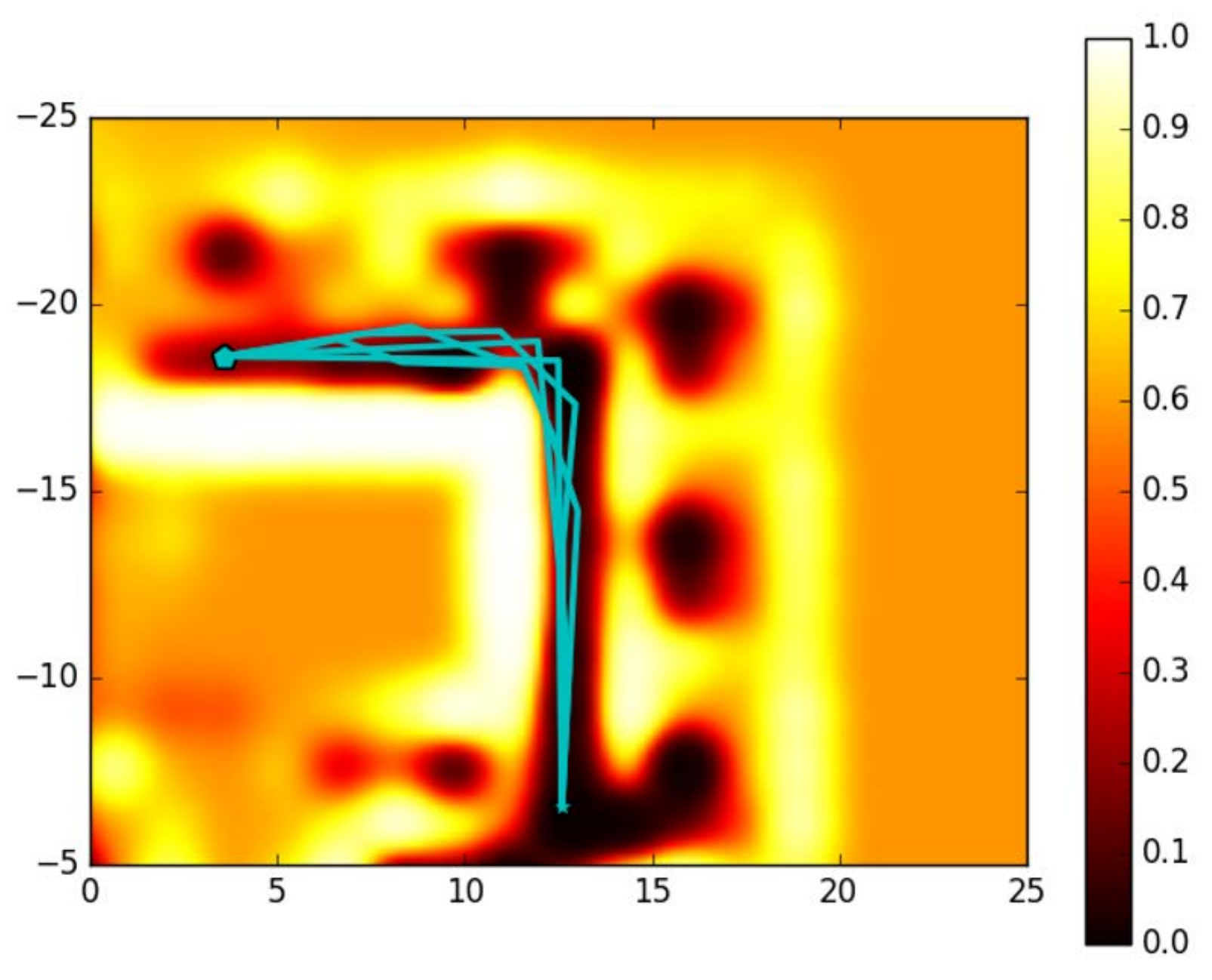}
 	}
     	\vfill
 	\subfloat[$\text{PRM}^*$ \label{Intel_Comparison:subfig-3}]{%
 		\includegraphics[width=0.23\textwidth]{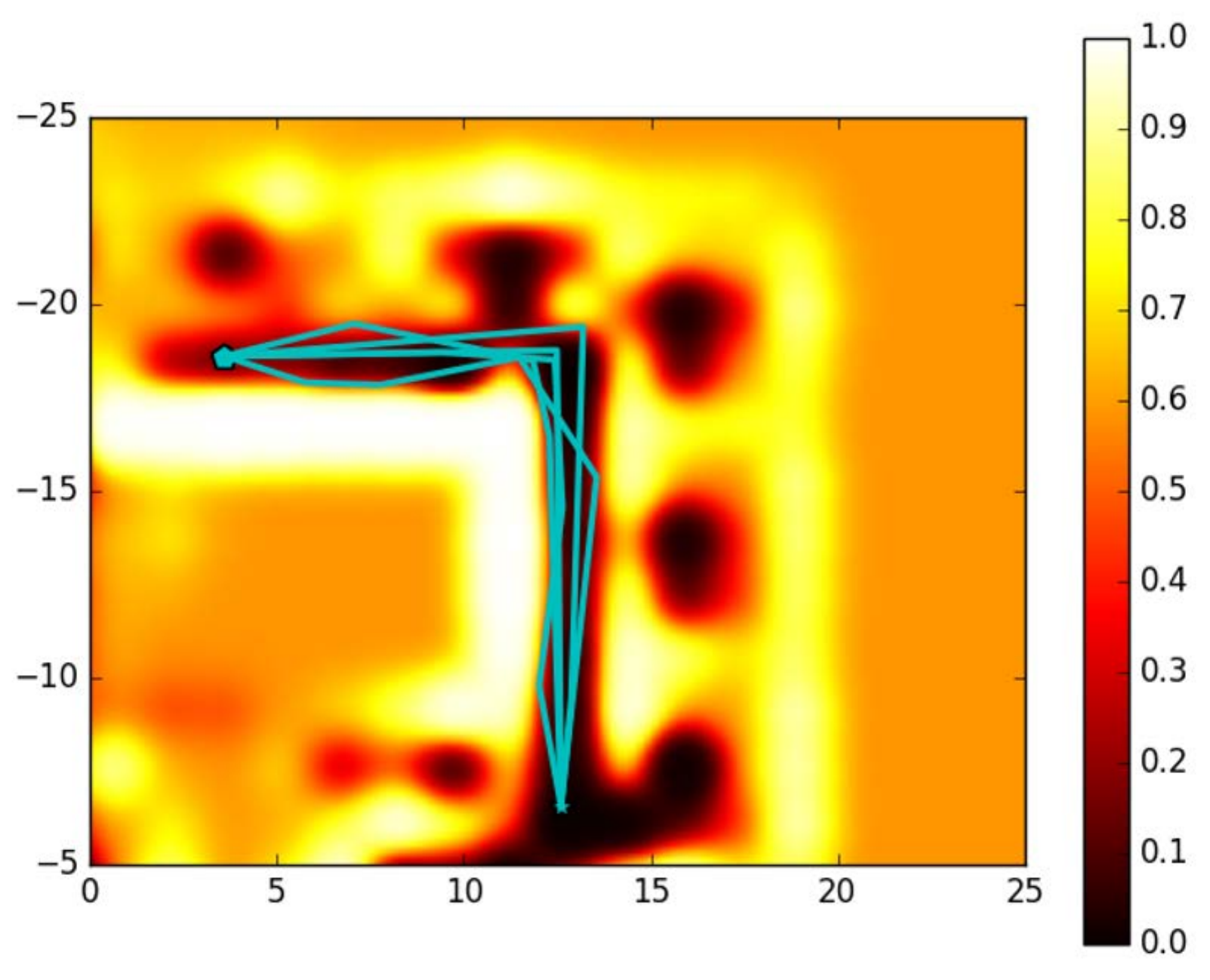}
     }   
    \hspace{1pt}
 	\subfloat[Convergence \label{Intel_Comparison:subfig-4}]{%
 		\includegraphics[width=0.23\textwidth]{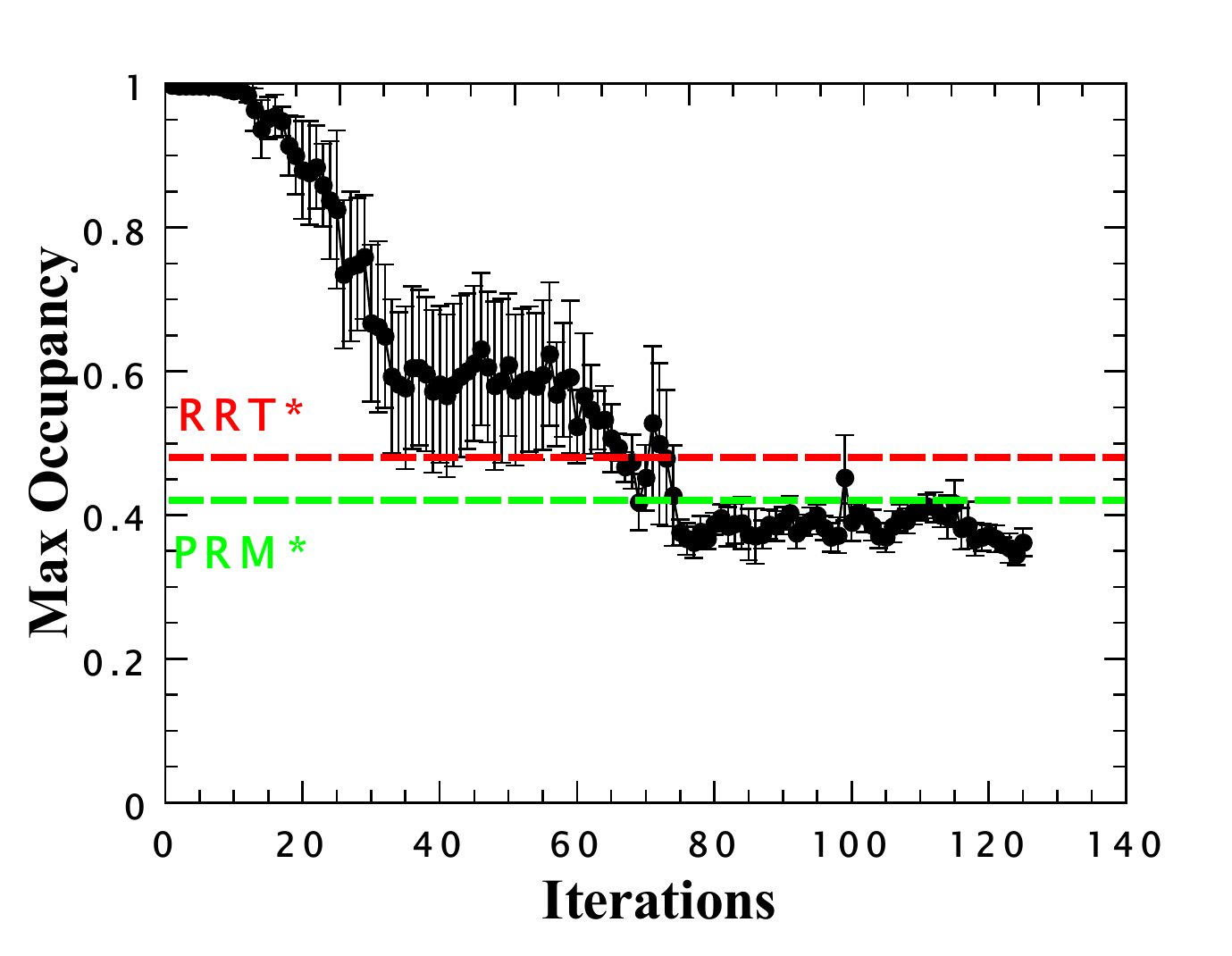}
 	}
 	\caption{Comparison of path planning methods on a continuous occupancy map of Intel-Lab; \protect\subref{Intel_Comparison:subfig-1} stochastic functional gradient motion planner (SFGMP), \protect\subref{Intel_Comparison:subfig-2} $\text{RRT}^*$ and \protect\subref{Intel_Comparison:subfig-3} $\text{PRM}^*$. Each image shows five repetitions of path planning with each method. SFGMP paths are smooth and follow the mid lines between walls. $\text{RRT}^*$ and $\text{PRM}^*$ produce paths that move the robot dangerously close to walls at times. \protect\subref{Intel_Comparison:subfig-4} shows convergence of the stochastic functional gradient motion planner. The maximum occupancy along the trajectory is plotted as a function of the iteration. Data shown is the average over 5 repetitions. Red and green dashed lines mark the average performance of $\text{RRT}^*$ and $\text{PRM}^*$ respectively. Our proposed stochastic planner significantly outperforms the other methods. Note that the maximum occupancy does not reduce to zero, as the the continuous occupancy map predictions for the end point is approximately 0.35.}
 	\label{fig:Intel_Comparison}
 \end{figure}

The map for this experiment was generated using the Intel-Lab dataset (available at http://radish.sourceforge.net/). This map contains many rooms and dead-ends that might challenge the optimiser. We compared the optimal trajectory of our proposed method with two other standard planning methods; $\text{RRT}^*$ \cite{karaman2010incremental} and $\text{PRM}^*$ \cite{Karaman2011} using implementations from the \textit{Open Motion Planning Library} (OMPL) \cite{sucan2012the-open-motion-planning-library}. Fig. \ref{fig:Intel_Comparison} shows a comparison between the different methods. Both $\text{RRT}^*$ and $\text{PRM}^*$ generate a path from start to goal. The path consists of waypoints (states) the robot should pass through. The list of waypoints provides a very sparse representation of the path that requires additional resources in order to transform into robot actions. In contrast, the proposed stochastic functional motion planner provides a detailed and smooth path represented by a function of $t$. Furthermore, the paths generated by $\text{RRT}^*$ and $\text{PRM}^*$ might follow close to walls or overshoot the corner leading to a higher risk of collision, and longer paths. The paths of our stochastic planner tend to follow the mid line between obstacles and perform smooth turns resulting in shorter and safer trajectories.  

Table \ref{tab:performance_comparison} and Fig. \ref{Intel_Comparison:subfig-4} provide a quantitative comparison between our stochastic functional gradient motion planner (SFGMP), $\text{RRT}^*$ and $\text{PRM}^*$. The objective of the optimisation is to minimise the obstacle cost. Fig. \ref{Intel_Comparison:subfig-4} shows the reduction of the maximum occupancy along trajectory as the optimisation progresses. After converging to the optimal solution, the safety of the path of the proposed method is significantly better than that of the other two methods. The results in Table \ref{tab:performance_comparison} summarises the expected performance. The maximum occupancy along the path, which indicates the path safety, is 0.36 for the proposed method and 0.42 and 0.48 for  $\text{PRM}^*$ and $\text{RRT}^*$, respectively. 
Furthermore, the penalty term, $\mathcal{U}_{dyn}$, in the objective in (\ref{eq:empirical_risk}) leads optimisation to prefer shorter paths. Consequently, the path generated by our method outperforms the other sampling-based methods.  
 \begin{table}[thbp]
	\centering
	\caption{Performance comparison}
	\label{tab:performance_comparison}
	\begin{tabular}{lrrr}
		\toprule
		& SFGMP & $\text{RRT}^*$ \cite{karaman2010incremental} & $\text{PRM}^*$ \cite{Karaman2011} \\
		\midrule
		Maximum occupancy & $ 0.36 \pm 0.02$ & $ 0.44 \pm 0.03$ & $ 0.46 \pm 0.03$ \\
		Path length [m]   & $20.90 \pm 0.10$ & $21.50 \pm 0.10$ & $22.50 \pm 0.50$ \\
        \bottomrule
	\end{tabular}
\end{table}

\section{Conclusions}
\label{sec:conclusions}

This paper introduced a novel method for path optimisation using occupancy maps. Sampling-based techniques are the prevalent method for path planning using occupancy maps. Although these techniques are flexible and have a high success rate in finding safe paths, optimising additional properties of the path such as length and execution time are not part of their reasoning. Trajectory optimisers, on the other hand, are designed for that purpose. Yet, the current implementations of trajectory optimisers require a finite resolution in the trajectory support and rely on having access to a well defined cost potential field. However, neither of these requirements are met by occupancy maps. Gradients obtained from the map's occupancy are noisy and not necessarily informative, which limits the choice of trajectory support, especially when the resolution is finite.

The planning method used in this paper employs stochastic optimisation to enhance the expressiveness of the basic functional gradient motion planner. It removes the need to commit to an a-priori parametric resolution, which allows our planner to better handle obstacles. The GP paths used in the planner provide a structured and flexible representation that can easily incorporate prior knowledge or initial solutions, such as coarse paths generated by a sampling based method. 

Future areas of work include improving convergence rates which could be approached by targeting under-sampled areas of the curve by biasing the stochastic sampling. Using the variance prediction provided by the GP path, one can employ a Bayesian optimiser to direct the sampling toward unexplored regions. Another avenue for improvement is to take advantage of modern multi-core systems by parallelising the optimisation.

\addtolength{\textheight}{-2cm}   





\bibliographystyle{IEEEtran}

\bibliography{ICRA2017}

\end{document}